\documentclass{SCIS2019}
\usepackage{amssymb}
\usepackage{lineno,hyperref}
\usepackage{CJK}
\usepackage{graphicx}
\usepackage{amsmath}
\usepackage{caption}
\usepackage{epstopdf}
\usepackage{wrapfig}
\usepackage{color}
\usepackage{float}
\usepackage{times}
\usepackage{amssymb,amsmath}
\usepackage{chngpage}
\usepackage{array}
\usepackage [latin1]{inputenc}

\modulolinenumbers[5]

\begin{document}

%%%%%%%%%%%%%%%%%%%%%%%%%%%%%%%%%%%%%%%%%%%%%%%%%%%%%%%
%%% Authors do not modify the information below
%%% ×÷Õß²»ÐèÒªÐÞ¸Ä´Ë´¦ÐÅÏ¢
\ArticleType{RESEARCH PAPER}
%\SpecialTopic{}
\Year{2019}
\Month{}
\Vol{61}
\No{}
\DOI{}
\ArtNo{}
\ReceiveDate{}
\ReviseDate{}
\AcceptDate{}
\OnlineDate{}
%%%%%%%%%%%%%%%%%%%%%%%%%%%%%%%%%%%%%%%%%%%%%%%%%%%%%%%

%%% title: ±êÌâ
%%%   \title{title}{title for citation}
\title{Dynamic Network Embedding via Incremental Skip-gram with Negative Sampling}{}

%%% Corresponding author: Í¨ÐÅ×÷Õß
%%%   \author[number]{Full name}{{email@xxx.com}}
%%% General author: Ò»°ã×÷Õß
%%%   \author[number]{Full name}{}
\author[1,2]{Hao Peng}{}
\author[1,2]{Jianxin Li}{{lijx@act.buaa.edu.cn}}
\author[1,2]{Hao Yan}{}
\author[2]{Qiran Gong}{}
\author[3]{Senzhang Wang}{}
\author[2]{\\Lin Liu}{}
\author[4]{Lihong Wang}{}
\author[5]{Xiang Ren}{}

%%% Author information for page head. Ò³Ã¼ÖÐµÄ×÷ÕßÐÅÏ¢
\AuthorMark{Hao Peng, et al}

%%% Authors for citation. Ê×Ò³ÒýÓÃÖÐµÄ×÷ÕßÐÅÏ¢
\AuthorCitation{Hao Peng, et al}

%%% Authors' contribution. Í¬µÈ¹±Ï×
%\contributions{Authors A and B have the same contribution to this work.}

%%% Address. µØÖ·
%%%   \address[number]{Affiliation, City {\rm Postcode}, Country}
\address[1]{Beijing Advanced Innovation Center for Big Data and Brain Computing, Beihang University, Beijing, China}
\address[2]{State Key Laboratory of Software Development Environment, Beihang University, Beijing, China}
\address[3]{Collage of Computer Science and Technology, Nanjing University of Aeronautics and Astronautics, Nanjing, China}
\address[4]{National Computer Network Emergency Response Technical Team/Coordination Center of China, Beijing, China}
\address[5]{Department of Computer Science, University of Southern California, Los Angeles, USA}

%%% Abstract. ÕªÒª
\abstract{Network representation learning, as an approach to learn low dimensional representations of vertices, has attracted considerable research attention recently.
It has been proven extremely useful in many machine learning tasks over large graph.
Most existing methods focus on learning the structural representations of vertices in a static network, but cannot guarantee an accurate and efficient embedding in a dynamic network scenario.
The fundamental problem of continuously capturing the dynamic properties in an efficient way for a dynamic network remains unsolved.
To address this issue, we present an efficient incremental skip-gram algorithm with negative sampling for dynamic network embedding, and provide a set of theoretical analyses to characterize the performance guarantee.
Specifically, we first partition a dynamic network into the updated, including addition/deletion of links and vertices, and the retained networks over time.
Then we factorize the objective function of network embedding into the added, vanished and retained parts of the network. Next we provide a new stochastic gradient-based method, guided by the partitions of the network, to update the nodes and the parameter vectors. The proposed algorithm is proven to yield an objective function value with a bounded difference to that of the original objective function.
The first order moment of the objective difference converges in order of $\mathbb{O}(\frac{1}{n^2})$, and the second order moment of the objective difference can be stabilized in order of $\mathbb{O}(1)$.
Experimental results show that our proposal can significantly reduce the training time while preserving the comparable performance.
We also demonstrate the correctness of the theoretical analysis and the practical usefulness of the dynamic network embedding.
We perform extensive experiments on multiple real-world large network datasets over multi-label classification and link prediction tasks to evaluate the effectiveness and efficiency of the proposed framework, and up to 22 times speedup has been achieved.}

%%% Keywords. ¹Ø¼ü´Ê

\maketitle

%%%%%%%%%%%%%%%%%%%%%%%%%%%%%%%%%%%%%%%%%%%%%%%%%%%%%%%
%%% The main text. ÕýÎÄ²¿·Ö
%%%%%%%%%%%%%%%%%%%%%%%%%%%%%%%%%%%%%%%%%%%%%%%%%%%%%%%
\vspace{-0.15in}
\section{Introduction}\vspace{-0.05in}
Recently network representation learning, also known as network embedding, has received considerable research attention.
That is due to the fact that many real-world problems in complex systems, such as recommended systems, social networks and  biology networks, etc, can be modelled as machine learning tasks over large network.
The idea of network embedding is to learn a mapping that projects each vertex in a network to a low dimensional and continuous distributed vector space, where each vertex is represented as a dense vector.
The mapping is learned with the objective of preserving the structural information of the original network in the geometric relationships among vertices' vector representations~\cite{Hamilton2017Representation}.
Network representation learning has been proven to be a useful tool for various real-world network mining tasks such as vertex community detection~\cite{Cavallari2017Learning}, recommended system~\cite{shi2018heterogeneous}, anomaly detection~\cite{hu2016embedding}, multi-label classification~\cite{Grover2016node2vec,Perozzi2014DeepWalk,Tang2015LINE,Tang2015PTE}, link prediction~\cite{Grover2016node2vec,Perozzi2014DeepWalk,Tang2015LINE,Man2016Predict}, knowledge representation~\cite{Ren2016Label}, etc.

Previous studies have proposed several prominent network embedding methods.
DeepWalk and node2vec capture higher-order proximities in embeddings by maximizing the conditional probability of observing the neighbourhood of vertices of a vertex given the mapped point of the vertex.
Here the neighbourhood vertices are obtained from vertices traversed in a random walk.
The crucial difference between DeepWalk~\cite{Perozzi2014DeepWalk} and node2vec~\cite{Grover2016node2vec} is that node2vec employs a biased random walk procedure to provide a trade-off between breadth-first search (BFS) and depth-first search (DFS) in a network, which might lead to a better mapping function.
LINE~\cite{Tang2015LINE} and SDNE~\cite{Wang2016Structural} learn graph embeddings by preserving the first- and second-order proximities in the embedded space, where the former refers to the pairwise neighborhood relationship and the latter is determined by the similarity of nodes' neighbors.
The difference is that the SDNE uses highly non-linear functions to represent the mapping function.

Most existing network embedding methods~\cite{cui2018survey} focus on learning the node representations in static network where no temporal information is associated with the nodes and edges.
However, the majority of real-world networks are dynamical and continuously growing over time (i.e., nodes occur and disappear, and edges are added and vanish as time goes), such as the friendship network in Facebook, the citation network in DBLP and the web-pages hyperlink dataset updating in Wikipedia, etc.
There are a lot of scenarios, such as real-time social network node classification and knowledge graph link processing, requiring dynamic update of the node representation given the fact that the working domains are fast evolving.
Unfortunately, the above methods ignore the dynamic nature and are unable to efficiently update the vertices' representations in accordance with networks' evolution.
However, prior works have demonstrated that, besides the dynamic edge and vertex modeling, the negative sampling or hierarchical softmax optimizing for representation learning is tremendous importance in capturing the evolution patterns of the dynamic network~\cite{Trivedi2017Know,Zuo2018}.
When the difference between the updated network and the old network is relatively small, it is inefficient to obtain the new node embeddings through retraining the entire new network.
Indeed, the few very recent works~\cite{Trivedi2017Know,Zhu2016Scalable,Zuo2018,Hamilton2017Inductive} adapted from the above methods require either prior knowledge of new vertices' attributes or retrain on new graphs with uncertain convergence time.
It's a challenge for many high-throughput production machine learning systems that need generating the representations of new vertices promptly.

In this paper, we study the problem of efficiently learning the node embedding for dynamic networks by proposing an incremental skip-gram with negative sampling model.
In particular, we adopt the popular and fundamental network representation models, such as DeepWalk and node2vec, due to their simplicity, interpretability, time efficiency, and comparable performance to other complex network embedding technologies~\cite{Chen2016Incorporate,Cao2015GraRep, Wang2016Structural, Tang2015LINE,Yang2017Fast}.
The two models make use of skip-gram which is initially proposed in natural language processing (NLP) to train vertice representations through generating the sequences of the vertex by random walk. To speed up the training process, unsupervised neural network based language learning models employ two techniques called hierarchical softmax and negative sampling~\cite{Mikolov2013Distributed, Mikolov2013Efficient}.
Hierarchical softmax was first proposed by Morin and Bengio~\cite{MorinB05} where a hierarchical tree is constructed to index all the words in a corpus as leaves. Negative sampling is developed based on noise contrastive estimation~\cite{GutmannH12} and randomly samples the words not in the context to distinguish the observed data from the artificially generated random noise.
The fixed number of the negative samples replaces the variable layers of hierarchy.
Although the original DeepWalk employs hierarchical softmax~\cite{Perozzi2014DeepWalk}, it can be also implemented using the negative sampling like node2vec, LINE, etc.
Considering the interpretability, popularity and good performance of skip-gram and negative sampling on various representation learning models~\cite{Grover2016node2vec, Perozzi2014DeepWalk, Tang2015LINE, Mikolov2013Distributed, Mikolov2013Efficient, Levy2014Neural,Yang2017Fast}, we investigate the problem of learning dynamic network embeddings with a focus on designing an Incremental Skip-Gram model with Negative Sampling (ISGNS).

\begin{figure}[h]
\centering
\includegraphics[width=0.8\textwidth,height=0.18\textheight]{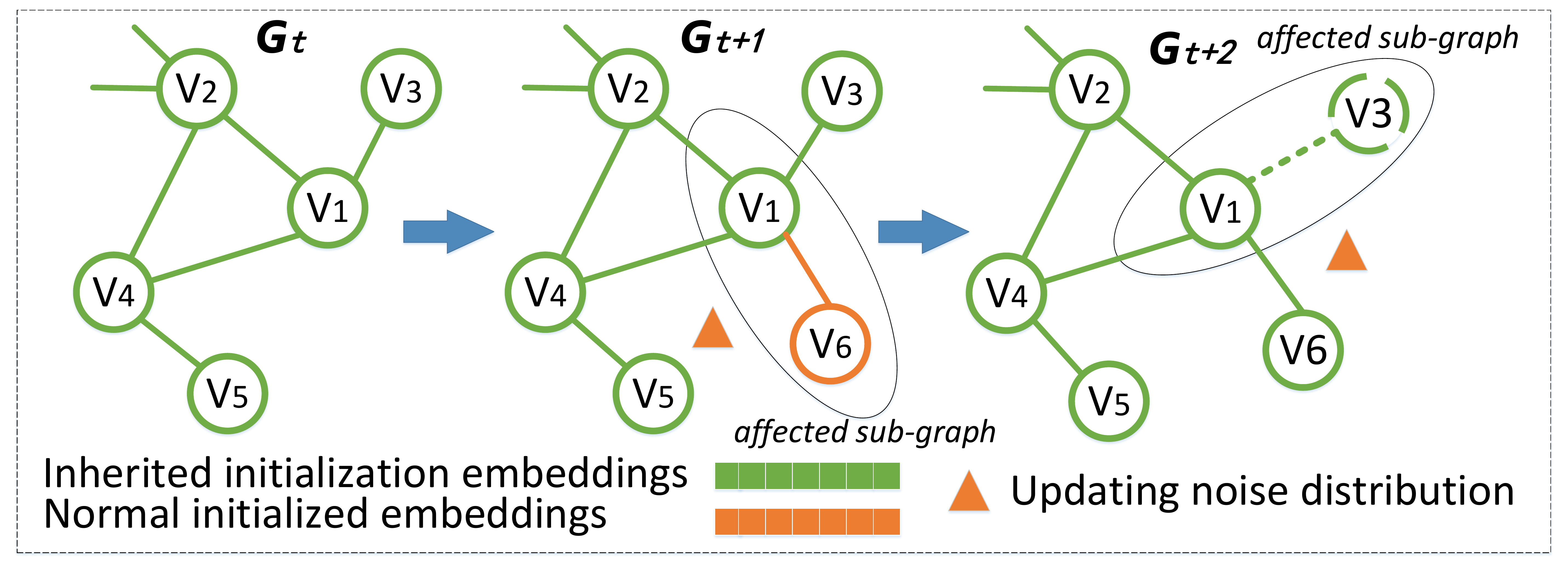}\vspace{-0.1in}
\caption{An illustration of the temporal evolving of the dynamic network.
The green vertices and edges constitute the initial network in time $t$.
The vertex $V_{6}$ (marked in orange color) and the corresponding edge $(V_{1},V_{6})$ (marked in orange color) emerge in time $t+1$.
The vertex $V_{3}$ (marked in dotted line) and the corresponding edge $(V_{1},V_{3})$ (marked in dotted line) vanish in time $t+2$.}\label{fig:dynamic}
\end{figure}

When applying skip-gram with negative sampling to network representation learning, the first problem is to investigate the structure proximities and compute the noise distributions for negative sampling~\cite{Grover2016node2vec, Perozzi2014DeepWalk, Tang2015LINE, Mikolov2013Distributed, Mikolov2013Efficient, Levy2014Neural,Yang2017Fast}.
When the vertices and edges of a network evolve over time, as shown in Figure~\ref{fig:dynamic}, the proximities and noise distributions will update automatically to reflect the change of the network structure.
For example, in DeepWalk and node2vec, they use edges to construct the sequences of the vertices and the noise distribution over the vocabulary, and result in the faster training process.
In the dynamic scenario, when the edges, the edge weights and the vertices change, the sequences of vertices, the structure proximities and the noise distribution should be updated correspondingly.
To address this issue, we first partition the network into the updated part (new added/vanished links and nodes) and the retained part.
Then, we employ random walk and sliding window~\cite{Perozzi2014DeepWalk, Grover2016node2vec} to extract the sequences of the nodes or subgraphs, namely affected sequences of subgraphs in the network.
To speed up model training, our model inherits all the retained nodes and parameter vectors and implements a new stochastic gradient-based method to update the changed nodes and parameter vectors, by comparing the old and updated networks.
When updating the vectors for the updated part of the subgraphs, we make use of stochastic gradient descent and ascent methods based on the latest noise distributions to optimize the model. In this way, we only need to update vectors in affected subgraphs.
Our theoretical analyses reveal that, under a mild assumption, the objective difference can be bounded by the scale of the old network, and the convergence of objective difference can also be bounded. So the optimal solution of the dynamic network embedding by ISGNS agrees with the original network SGNS when the network scale is infinitely large. Since the update process is independent of all the shared vectors, we also present the techniques for an efficient parallel implementation of dynamic network embedding with ISGNS.
In the experiments, we show that the proposed model can significantly reduce the training time while preserving comparable performances with state-of-the-art models on static networks.
The code of this work is publicly available at \url{https://github.com/RingBDStack/dynamic_network_embedding}.

Our main contributions are summarized as follows.
\begin{itemize}
\item A dynamic network embedding framework based on an approximately optimal solution of incremental skip-gram with negative sampling is proposed, which can be directly applied in existing network embedding models such as DeepWalk and node2vec.
\item The solid theoretical analyses show that our proposal guarantees the boundness of the objective difference and the convergence when the training network scale is infinitely large. The empirical study also verifies the boundary and moments of the network dynamic change.
\item Extensive experiments on multiple large real-world network datasets show both the efficiency and effectiveness of the proposed ISGNS on multi-label classification and link prediction tasks.  ISGNS achieves up to 22 times speedup while preserving comparable performance with global re-training methods.
\end{itemize}

The remainder of the paper is organized as follows.
We first review the related work in Section~\ref{sec:relatedwork}.
Then we introduce the proposed ISGNS model in detail in Section~\ref{sec:DNE}.
Section~\ref{sec:theory} provides the mathematical details of the dynamic objective difference, corresponding bound analysis and convergence analysis for both first-order and second-order moments.
We evaluate our model in Section~\ref{sec:experiments}, and
finally conclude this work in Section~\ref{sec:conclusion}.

\vspace{-0.15in}
\section{Related Work}\label{sec:relatedwork}\vspace{-0.05in}
In this section, we briefly review related work on network embedding models, including static network embedding and dynamic network embedding technologies.

\textbf{Static network embedding}.
DeepWalk~\cite{Perozzi2014DeepWalk} is the first work that utilizes a truncated random walk to transform a static network into a collection of node sequences.
Then the skip-gram on hierarchical softmax function\footnote{However, an alternative to the hierarchical softmax is Noise Contrastive Estimation (NCE)~\cite{Gutmann2012Noise,Mnih2012A}} is utilized to learn the vertex representations.
Node2vec~\cite{Grover2016node2vec} further generalizes DeepWalk with Breadth-First Search (BFS) and Depth-First Search (DFS) on random walks, and employs the popular skip-gram with negative sampling to learn the vertex representations.
LINE~\cite{Tang2015LINE} and SDNE~\cite{Wang2016Structural} model the first-order and second-order proximities between vertices, and employ the skip-gram with negative sampling to deal with the limitation of stochastic gradient descent on weighted edges without compromising the efficiency.
Struct2vec~\cite{Ribeiro2017struc2vec} proposes to preserve the structural identity between nodes in the representation.
To achieve this goal, it first creates a new graph based on the structural identity similarity between nodes and then follows a similar method to DeepWalk on the created graph.
A very recent method Graph-Wave~\cite{Donnat2018Learning} makes use of wavelet diffusion patterns by treating the wavelets from the heat wavelet diffusion process as distributions.
Overall, those methods of generalized network embeddings are typically designed to go through the entire network multiple times.
It means that they cannot perform online learning of the node representation in a dynamic scenario when the vertices, edges and edge weights change over time.

\textbf{Dynamic network embedding}.
DANE~\cite{Li2017Attributed} leverages a matrix perturbation theory to update the dynamic attributed network spectral embeddings.
Zhu et al.~\cite{Zhu2016Scalable} proposes a temporal latent space learning model BCGD via non-negative matrix factorization to target the link prediction task in dynamic social networks.
But it belongs to especial embedding method for the purpose of link prediction.
Trivedi et al.~\cite{Trivedi2017Know} proposes a deep recurrent architecture Know-Evolve modeling the historical evolution of entity representations in a specific relationship space.
Xu et al.~\cite{Xu2013Dynamic} proposes a statistical model Dynamic SBM for dynamic networks that utilized a set of unobserved time-varying states to characterize the dynamics of the network.
Jian et al.~\cite{Jian2018Toward} designs an online embedding representation learning method OLSN based on spectral embedding used for node classification.
Zhou et al.~\cite{zhou2018dynamic} proposes a triadic closure process based semi-supervised algorithm Dynamic Triad to learn the structural information and evolution pattern in dynamic networks.
Du et al.~\cite{du2018dynamic} proposes a heuristic dynamic network embedding method DNE, which employs a decomposable objective based on the skip-gram objective, and give the objective function difference minimization.
Zuo et al.~\cite{Zuo2018} proposes a Hawkes process based temporal network embedding method HTNE which captures the influence of the historical neighbors on the current neighbor formation simultaneously.
Inspired by the unsupervised neural network representation learning, previous incremental word embedding models~\cite{peng2017incrementally,Kaji2017Incremental,Rudolph2018Dynamic,peng2018incremental} proposed the incremental hierarchical softmax function, the small adaptive unigram table based negative sampling for incremental word embeddings, and Gaussian random walk based dynamic word embedding.
For existing dynamic network embedding and analysis models~\cite{Li2017Attributed,Zhu2016Scalable,Trivedi2017Know,Xu2013Dynamic,Jian2018Toward,Zuo2018,du2018dynamic}, these models belong to heuristic methods, and cannot theoretically guarantee the equivalence and optimality of generalized network embedding objective function.
Even, the computational cost or memory cost linearly increases with the assumes and learning time.
For the bright dynamic network embedding model~\cite{zhou2018dynamic}, it can not handle the addition of vertices, and the scalability of the model and the hypothetical process are the bottlenecks when applied in real large scale networks.
In addition, the above discussed dynamic network representation learning models have not strictly followed the original object in the sampling optimization.
Therefore, different from existing work, we study the popular neural network based dynamic network embedding from the perspectives of the objective function and the sampling strategy.

\vspace{-0.15in}
\section{Dynamic Network Representation Learning}\label{sec:DNE}\vspace{-0.05in}
As we discussed above, network representation learning is sensitive to the network structure and the objective proximities among vertices.
When the edges and vertices evolve over time, we depict the structural and proximity differences of the network snapshots in different time slots by metabolic sub-graphs, which directly reflect the changes in edges, vertices, and noise distributions.
Inspired by the principle of network embedding approximating the adjacency matrix~\cite{Yang2017Fast}, we assume that the influence of structural changes on the representation learning is partial in neighborhoods/sub-graphs for limited adjacency matrix float.
We firstly locate the metabolic sub-graphs by local random walk in re-training.
Then, for newly added nodes or edges, we implement the random walk only on the sub-graphs to generate sequences of vertices.
Note that one vertex sequence contains at least one new node or edge.
For the vanished nodes or edges, we also implement the random walk on the sub-graphs to generate sequences of vertices following the same rule.
Then, we re-count the frequency for each vertex appearing in the above sequences, and add/subtract its into/from the frequencies generated in the old network.
Thus, we obtain the latest noise distributions.
For a fast training in dynamic network scenario, we adopt a strategy that inherits the vertexes and the parameter vectors through changes in the network structures.
If part of the network remains the same, we can retain the vectors as well as the structure associated to the nodes, and distinguish the updated sub-graphs between the old network and the new network structures.

\subsection{Node Initialization and Inheritance}\label{sec:nodeinitiainheri}
Given a network $\mathcal{W}$ in time $t$, we formulate the updated network $\mathcal{W}'$ in time $t+1$ as:
\begin{equation}\label{eq:dynamicnetwork}\vspace{-0.1in}
  \mathcal{W}'=\mathcal{W} + \Delta \mathcal{W}_{inc}- \Delta \mathcal{W}_{dis}\quad,
\end{equation}
where $\Delta \mathcal{W}_{inc}$ and $\Delta \mathcal{W}_{dis}$ refer to the newly added and vanished sub-graphs, respectively.
We re-calculate the noise distributions for each vertex on new network $\mathcal{W}'$ in time $t+1$ with random walk \cite{Grover2016node2vec,Perozzi2014DeepWalk} on the above sub-graphs.

We first preserve all the vertices and the corresponding parameter vectors in the old network.
Then we inherit the reserved vertex and parameter vectors as initialization in the new network.
If a vertex is newly added, we initialize it as a random vector with the same dimension as the existing vertexes, and the related parameter vector is initialized as a zero vector.
It can be formally defined as follows:
\begin{equation}\label{eq:parametersinheritance1}
{v}'(u)=
\begin{cases}
v(u), &\mbox{$u\in \mathcal{W}$}\\
{\rm random}, &\mbox{$u\notin \mathcal{W}$}
\end{cases}\quad,
\end{equation}
and
\begin{equation}\label{eq:parametersinheritance2}
\widetilde{v}'(u)=
\begin{cases}
\widetilde{v}(u), &\mbox{$u\in \mathcal{W}$}\\
{\rm \bm{0} }, &\mbox{$u\notin \mathcal{W}$}
\end{cases}\quad,
\end{equation}
where ${v}(u)$ and ${v}'(u)$ are the representation vectors of node $u$ for the old and new networks, and $\widetilde{v}(u)$ and $\widetilde{v}'(u)$ are the parameter vectors of $u$ in the old and new networks, respectively.

\subsection{Model Updating}\label{sec:model-update}
In a dynamic network, we assume the updated nodes and edges only affect the representations of the local nodes and edges, and perform an approximate stochastic gradient method to update the related vectors.
In detail, after generating the sequence of vertices, given the size of sliding window $2c$ , we can build local sub-graphs, named as affected sub-graphs, for the newly added and vanished vertices and edges.
The approximate stochastic gradient method can be described as following two steps.
First, for the vanished vertices and edges, we perform a stochastic gradient descent method to update the old network representations with the updated noise distributions.
Second, after inheriting and initializing the vertexes and the parameter vectors, for the newly added vertexes and edges, we perform a stochastic gradient ascent method to update the new network representations.
More specific, we extend the widely used skip-gram with negative sampling method to dynamic sampling scenarios for network embeddings.
In terms of vertex representation updating, this yields to such an optimization problem:
\begin{equation}\label{eq:objective-frac1}
\mathop{max}\limits_{f}\quad \sum_{u\in\mathcal{W}'}log\quad Pr(N_{S'}(u)|f(u))\quad,
\end{equation}
where $f$ is the mapping function from the nodes to the feature representations, and $N_{S'}(u)$ refers to the neighborhood or context nodes of node $u$ generated through a sampling optimization strategy $S'$.

We aim to optimize the above objective function,
which maximizes the log-probability of observing a network partitioning.
The objective in Eq.~(\ref{eq:objective-frac1}) can be approximatively simplified to:
\begin{equation}\label{eq:objective-frac2}
\begin{aligned}
&\mathop{max}\limits_{f}\quad \sum_{u\in \mathcal{W}}[-log Z_{u} + \sum_{n_{i}\in N_{S(u)}}f(n_{i})\cdot f(u)]+&(\sum_{u\in \Delta \mathcal{W}_{inc}}-\sum_{u\in \Delta \mathcal{W}_{dis}})[-log Z_{u} + \sum_{n_{i}\in N_{S'(u)}}f(n_{i})\cdot f(u)]\quad,
\end{aligned}
\end{equation}
where $Z_{u} = \sum_{u\in\mathcal{W}'}exp(f(u)\cdot f(v))$ is expensive to compute for dynamic and large networks.
So we approximate it by negative sampling.
$S$ is the sampling optimization strategy in the old network $\mathcal{W}$. We factorize log-likelihood function for skip-gram with negative sampling model based on the network partitioning.
Here, we firstly retain the log-likelihood function and the inherited vectors from the old network.
Second, we re-calculate the sampling optimization strategy $S'$.
Then we employ the sub-graph compensation strategy to increase or decrease the holistic log-likelihood function, respectively.
Note that the difference between Eq.~(\ref{eq:objective-frac1}) and Eq.~(\ref{eq:objective-frac2}) is that the sampling optimization strategies $S$ and $S'$ are different.

Our goal is to speed up the training of dynamic network representation learning.
To train the node representations for the updated parts of a network, existing methods need to re-scan and re-train the whole proximities based on skip-gram with negative sampling and stochastic gradient methods.
Given the above factorization analysis of the objective function, we found that for the old network $\mathcal{W}$, we can apply the initialization and inheritance of vertices and parameters trick in Eq.~(\ref{eq:parametersinheritance1}) and Eq.~(\ref{eq:parametersinheritance2}) to significantly save the training time.
We just need to update the related vertex and parameter vectors following the new sampling optimization strategy $S'$.
Finally, we release the disappearance of nodes and the corresponding parameters vector.

\vspace{-0.15in}
\section{Theoretical Analysis}\label{sec:theory}\vspace{-0.05in}
Although the extension from the batch global training to the dynamic network training is simple and intuitive, it is not clear whether the incremental skip-gram with negative sampling technology based dynamic network embedding method can learn the represented vectors of the nodes as good as that learned by the batch global learning counterpart.
To answer this question, in this section we examine the dynamic network representation learning from a theoretical point of view.

We will firstly show the difference between the objectives optimized by the approximative incremental skip-gram with negative sampling (ISGNS) and batch skip-gram with negative sampling (SGNS) models in Section~\ref{sec:theory-objectivediff}.
Second, we will prove that the objective difference is bounded by the scale of the network in Section~\ref{sec:theory-boundness}.
Then, we will investigate the probabilistic properties of the objective difference to demonstrate the equivalent relationship between batch SGNS and ISGNS in Section~\ref{sec:theory-convergence}.
Finally, we will analyze the time and memory complexity of ISGNS in Section~\ref{sec:complexity}.

\subsection{Objective Difference}\label{sec:theory-objectivediff}
As discussed in Section~\ref{sec:nodeinitiainheri}, the network updates from $\mathcal{W}$ to $\mathcal{W}'$ and the size of vertex sequences growing from $n$ to $N$.
We denote the size of the local sequences that contain the newly added and vanished vertices as $n_{add}$ and $n_{van}$.
Following the works~\cite{Grover2016node2vec,Perozzi2014DeepWalk}, the size of vertex sequences can be updated as $N=n + n_{add} - n_{van}$.
The ISGNS optimizes the following objective function:
\begin{equation}\label{eq:DSGNS}
\begin{aligned}
\centering
\mathcal{L}_{{\rm ISGNS}}(\theta) = &-\{\frac{1}{n}\sum_{i=1}^{n}\sum_{|j|<c,j\neq0}\psi_{w_{i},w_{i+j}}^{+}+k\mathbb{E}_{v\sim q_{n}(v)}[\psi_{w_{i},v}^{-}]\\
&+(\frac{1}{n_{add}}-\frac{1}{n_{van}})\sum_{i=1}^{N}\sum_{|j|<c,j\neq0}\psi_{w_{i},w_{i+j}}^{+}+k\mathbb{E}_{v\sim q_{N}(v)}[\psi_{w_{i},v}^{-}]\}\quad,
\end{aligned}
\end{equation}
where $\theta =(\bm{t}_{1},\bm{t}_{2},\cdots,\bm{t}_{\mathcal{W}'},\bm{c}_{1},\bm{c}_{2},\cdots,\bm{c}_{\mathcal{W}'})$ collectively represents model parameters, including both target and context vertex embeddings.
The function $q_{n}(v)$ represents the old noise distribution, and it is defined as $q_{n}(v) = \frac{f_{n}(v)^{\frac{3}{4}}}{\sum_{v'\in {\mathcal{W}}} f_{n}(v')^{\frac{3}{4}}}$, where $f_{n}(v)$ represents the frequency of vertex $v$ in the sequences of vertices.
Note that the noise distribution in the first term of the objective is $q_{n}(v)$ rather than $q_{N}(v)$.
Because we employ the parameter initialization strategy and it can be seen as a simple approximation of the gradient.
In detail, $\psi_{w,v}^{+}=log\sigma(t_{w}\cdot c_{v})$, $\psi_{w,v}^{-}=log\sigma(-t_{w}\cdot c_{v})$, and $\sigma(x)$ is the sigmoid function.
Given a target-context vertex pair ($w_{i},w_{i+j}$), and $k$ negative samples $(v_{1},v_{2},\cdots,v_{k})$ sampled from the latest noise distribution $q_{N}(v)$, the gradient of $-\psi_{w_{i},w_{i+j}}^{+}-k\mathbb{E}_{v\sim q_{N}(v)}[\psi_{w_{i},v}^{-}]$ is computed at each step.

In contrast, the original objective function of re-training the network embedding model based on SGNS can be given as:
\begin{equation}\label{eq:SGNS}
\begin{aligned}
\centering
\mathcal{L}_{{\rm SGNS}}(\theta) = -\frac{1}{N}\sum_{i=1}^{N}\sum_{|j|<c,j\neq0}\psi_{w_{i},w_{i+j}}^{+}+k\mathbb{E}_{v\sim q_{N}(v)}[\psi_{w_{i},v}^{-}],
\end{aligned}
\end{equation}
which can be interpreted as the re-training procedure with SGD.
Since the expectation terms in the objectives can be rewritten as $\mathbb{E}_{v\sim q_{N}(v)}[\psi_{w_{i},v}^{-}]=\sum_{v\in \mathcal{W'}}q_{N}(v)\psi_{w_{i},v}^{-}$, the difference between the two objectives can be formalized as follows:
\begin{equation}\label{eq:Sum-Diff}
\begin{aligned}
\Delta\mathcal{L}_{{\rm DI}}(\theta) =& \mathcal{L}_{{\rm SGNS}}(\theta) -\mathcal{L}_{{\rm ISGNS}}(\theta) =\frac{1}{n}\sum_{i=1}^{n}\sum_{|j|<c,j\neq0}k\sum_{v\in \mathcal{W}'}(q_{N}(v)-q_{n}(v))\psi_{w_{i},v}^{-}\\
=&\frac{2ck}{n}\sum_{i=1}^{n}\sum_{w,v\in \mathcal{W}'}\delta_{w_{i},w}(q_{N}(v)-q_{n}(v))\psi_{w,v}^{-}\quad,
\end{aligned}
\end{equation}
where $\delta$ is the delta function.

\subsection{Boundness Analysis of $\Delta\mathcal{L}_{{\rm DI}}(\theta)$}\label{sec:theory-boundness}
To verify the correctness of our dynamic network embedding framework, we present the boundness analysis of the objective difference $\Delta\mathcal{L}_{{\rm DI}}(\theta)$ in this subsection.

We first give a theorem as follows.

\textbf{Theorem 1}.
\emph{The objective difference $\Delta\mathcal{L}_{{\rm DI}}(\theta)$ can be directly bounded by the scale of the old network as following:}
\begin{equation}\label{eq:Bound-Diff}
\begin{aligned}
\centering
\Delta\mathcal{L}_{{\rm DI}}(\theta)  < \frac{2ck}{n}\epsilon = \frac{2ck}{N-(\dot{n}-\ddot{n})}\epsilon.
\end{aligned}
\end{equation}

\emph{Sketch of Proof}.
The delta function can be loosely considered as a function on the real line which is zero everywhere except at the origin, where it is infinite
\begin{equation}\label{eq:delta}
\delta_{w_{i},w}(q_{N}(v)-q_{n}(v)) =
\begin{cases}
+\infty, &q_{N}(v)\neq q_{n}(v)\\
0, &q_{N}(v)=q_{n}(v)
\end{cases}\quad ,
\vspace{-0.05in}
\end{equation}
and it is also constrained to satisfy the identity
\begin{equation}\label{eq:Delta-condition}
\begin{aligned}
\centering
\sum_{w,v\in \mathcal{W}'}\delta_{w_{i},w}(q_{N}(v)-q_{n}(v)) \leqslant 1.
\end{aligned}
\end{equation}
Since $\psi_{w,v}^{-}=log\sigma(-t_{w}\cdot c_{v})$ is bounded in practice,  we assume $\psi_{w,v}^{-} < \epsilon$.
It supports the very intuitive understanding that the less updated network nodes lead to a lower upper bound.

\subsection{Convergence Analysis of $\Delta\mathcal{L}_{{\rm DI}}(\theta)$}\label{sec:theory-convergence}
It shows that the first order of $\Delta\mathcal{L}_{{\rm DI}}(\theta)$ has an analytical form.

\textbf{Definition 1}.
Let $X_{i,w}$ be a random variable that represents $\delta_{w_{i},w}$.
It is assigned value 1 when the $i$-th node in the sampled data is $w \in \mathcal{W}'$.
For any $i$ and $j$, remind that $\mathbb{E}[X_{i,w}] = \mu_{w}$ and $\mathbb{V}[X_{i,w},X_{j',w}] = \rho_{w,v}$.

\textbf{Definition 2}.
Let $Y_{i,w}$ be a random variable that represents $q_{N}(v)$.

\textbf{Theorem 2}.
\emph{The first-order moment of $\Delta\mathcal{L}_{{\rm DI}}(\theta)$ is given as}
\begin{equation}\label{eq:Expection-Diff-SGNS}
\begin{aligned}
\centering
\mathbb{E}[\Delta\mathcal{L}_{{\rm DI}}(\theta)] =\frac{2ck}{n}(\frac{1}{N}-\frac{1}{n})\sum_{w,v\in \mathcal{W}'}\rho_{w,v}\psi_{w,v}^{-}\quad,
\end{aligned}
\end{equation}
where $\rho_{w,v}$ is the covariance of $X_{i,w}$ and $X_{j,v}$.

\emph{Sketch of Proof}.
Here, for any $i$ and $j$ such that $i<j$, we have
\begin{equation}\label{eq:Sub-Expection}
\begin{aligned}
\centering
\mathbb{E}[X_{i,w}Y_{j,v}] &= \mathbb{E}[X_{i,w}\frac{1}{j}\sum_{j'=1}^{j}X_{j',v}]=\frac{1}{j}\sum_{j'=1}^{j}\mathbb{E}[X_{i,w}X_{j',w}] = \frac{1}{j}\sum_{j'=1}^{j}(\mathbb{E}[X_{i,w}]\mathbb{E}[X_{j',v}]+\mathbb{V}[X_{i,w},X_{j',w}])\\
&=\mu_{w}\mu_{v}+\frac{1}{j}\rho_{w,v}\quad.
\end{aligned}
\end{equation}
Therefore, $\mathbb{E}[\Delta\mathcal{L}_{{\rm DI}}(\theta)] $ can be written as
\begin{equation}\label{eq:Expect-DiffSGNS}
\begin{aligned}
\centering
\mathbb{E}[\Delta\mathcal{L}_{{\rm DI}}(\theta)]
= & \frac{2ck}{n}\sum_{w,v\in \mathcal{W}'}(\mu_{w}\mu_{v}+\frac{1}{N}\rho_{w,v}-\mu_{w}\mu_{v}-\frac{1}{n}\rho_{w,v})\psi_{w_{r},v}^{-} =  \frac{2ck}{n}(\frac{1}{N}-\frac{1}{n})\sum_{w,v\in \mathcal{W}'}\rho_{w,v}\psi_{w,v}^{-}.
\end{aligned}
\end{equation}

\textbf{Theorem 3}.
\emph{The first-order moment of $\Delta\mathcal{L}_{{\rm DI}}(\theta)$ decreases in the order of $\mathcal{O}(\frac{1}{n^2})$}:
\begin{equation}\label{eq:Estimation1}
\begin{aligned}
\centering
\mathbb{E}[\Delta\mathcal{L}_{{\rm DI}}(\theta)] = \mathcal{O}(\frac{1}{n^2})\quad,
\end{aligned}
\end{equation}
\emph{and thus converges to zero in the limit of infinity:}
\begin{equation}\label{eq:Estimation-Euler-Harmonic1}
\begin{aligned}
\centering
{\lim\limits_{n\to\infty}}\mathbb{E}[\Delta\mathcal{L}_{{\rm DI}}(\theta)] = 0.
\end{aligned}
\end{equation}
\emph{Proof.} We assume that $N$ and $n$ are in the same order of magnitude and thus Theorem 2 gives the proof.

\textbf{Theorem 4}.
\emph{The second-order moment of $\Delta\mathcal{L}_{{\rm DI}}(\theta)$ can be bounded as}
\begin{equation}\label{eq:2Diff-SGNS}
\begin{aligned}
\centering
\mathbb{E}[\Delta\mathcal{L}_{\rm DI}^2(\theta)] < \sum_{w,v \in \mathcal{W}'}[\frac{24c^{2}k^{2}}{L^{2}T^{2}} + \mathcal{O}(\frac{1}{n})](\psi_{w,v}^{-})^2.
\end{aligned}
\end{equation}
\emph{and thus decreases in the order of $\mathcal{O}(1)$:}
\begin{equation}\label{eq:2Estimation}
\begin{aligned}
\centering
{\lim\limits_{n\to\infty}}\mathbb{E}[\Delta\mathcal{L}_{\rm DI}^2(\theta)] = \mathcal{O}(1),
\end{aligned}
\end{equation}
where $L$ refers to the random walk steps in one round, and $T$ regers to the random walk times for each vertex.

\emph{Proof}.
A similar result to first-order moment of $\Delta\mathcal{L}_{{\rm DI}}(\theta)$  can be proved for the second order moment of objective difference as well.
The upper-bound of $\mathbb{E}[\Delta\mathcal{L}_{\rm DI}^2(\theta)]$ is examined to prove the theorem.
Let $\Psi_{i,N,n,w,v}=\delta_{w_{i},w}(q_{N}(v)-q_{n}(v))\psi_{w,v}^{-}$.
Making use of Jensen's inequality, we have
\begin{equation}\label{eq:2Diff-SGNS}
\begin{aligned}
\centering
\mathbb{E}[\Delta\mathcal{L}_{{\rm DI}}^2(\theta)] =&\mathbb{E}[\frac{4c^{2}k^{2}}{n^{2}}(\sum_{i=1}^{n}\sum_{w,v\in \mathcal{W}'}\Psi_{i,N,n,w,v})^2] = \mathbb{E}[\frac{4c^{2}k^{2}}{n^{2}}|\mathcal{W}'|^{4}n^{2}(\sum_{w,v\in \mathcal{W}'}\sum_{i=1}^{n}\frac{1}{|\mathcal{W}'|^{2}n}\Psi_{i,N,n,w,v})^2]\\
\leqslant & \mathbb{E}[\frac{4c^{2}k^{2}}{n^{2}}|\mathcal{W}'|^{4}n^{2}\sum_{w,v\in \mathcal{W}'}\sum_{i=1}^{n}\frac{1}{|\mathcal{W}'|^{2}n}\Psi_{i,N,n,w,v}^2] = \frac{4c^{2}k^{2}|\mathcal{W}'|^2}{n}\sum_{w,v\in \mathcal{W}'}\sum_{i=1}^{N}\mathbb{E}[\Psi_{i,N,n,w,v}^2].
\end{aligned}
\end{equation}

To prove Theorem 4, we begin by examining the upper- and lower-bounds of $\mathbb{E}[X_{i,w}Y_{j,v}Y_{k,v}]$ in the
following Lemma, and then make use of the bounds to evaluate the order of the second order moment of $\Delta\mathcal{L}_{{\rm DI}}(\theta)$.

\textbf{Lemma 1}. For any $j$ and $k$ such that $j\le k$, we have
\begin{equation}\label{eq:2Diff-SGNS}
\begin{aligned}
\centering
\mathbb{E}[X_{i,w}Y_{j,v}Y_{k,v}]&\leqslant\frac{(jk-2j-k+2)\mu_{w}\mu_{v}^2+2j+k-2}{jk},\\
\mathbb{E}[X_{i,w}Y_{j,v}Y_{k,v}]&\geqslant\frac{(jk-2j-k+2)\mu_{w}\mu_{v}^2}{jk}.
\end{aligned}
\end{equation}
See the Supplementary File for detailed proof.
Furthermore, the term $\mathbb{E}[\Delta\mathcal{L}_{{\rm DI}}^2(\theta)]$ is upper-bounded as
\begin{equation}\label{eq:2Estimation-SGNS}
\begin{aligned}
\centering
&\mathbb{E}[\Psi_{i,N,n,w,v}^2]=\mathbb{E}[\delta_{w_{i},w}(q_{N}(v)-q_{n}(v))^2(\psi_{w,v}^{-})^2]
 %= \mathbb{E}[X_{i,w}(Y_{N,v}-Y_{n,v})^2](\psi_{w,v}^{-})^2\\
%=&(\mathbb{E}[X_{i,w}Y_{N,v}^{2}]-2\mathbb{E}[X_{i,w}Y_{N,v}Y_{n,v}] + \mathbb{E}[X_{i,w}Y_{n,v}^{2}])(\psi_{w,v}^{-})^2\\
%\leqslant & \{-\frac{1}{N}\mu_{w}\mu_{v}^2+\frac{2}{N^2}\mu_{w}\mu_{v}^2-\frac{1}{n}\mu_{w}\mu_{v}^2+\frac{2}{n^2}\mu_{w}\mu_{v}^2 - \frac{4}{nN}\mu_{w}\mu_{v}^2+\frac{3}{N}-\frac{2}{N^2}+\frac{3}{n}-\frac{2}{n^2}\}(\psi_{w,v}^{-})^2\\
< &\sum_{w,v \in \mathcal{W}}[\frac{3}{N}+\frac{3}{n}+(\frac{2}{N^2}+\frac{2}{n^2})\mu_{w}\mu_{v}^2](\psi_{w,v}^{-})^2.
\end{aligned}
\end{equation}

Since the sequence of vertices are generated by random walk technologies, the mathematical relationship between set of vertices $\mathcal{W}'$ and set of sequences $N$ can be formalized as
\begin{equation}\label{eq:vertices-sequences}
\begin{aligned}
\centering
N = \mathcal{W}'\cdot L\cdot T.
\end{aligned}
\end{equation}
%where $L$ refers to random walk steps in one round, and $T$ to be random walk times for each vertex in practice.
So, the upper-bounded of $\mathbb{E}[\Delta\mathcal{L}_{\rm DI}^2(\theta)]$ can be written as:
\begin{equation}\label{eq:2Diff-SGNS}
\begin{aligned}
\centering
\mathbb{E}[\Delta\mathcal{L}_{\rm DI}^2(\theta)]
<& \sum_{w,v \in \mathcal{W}'} \frac{4c^{2}k^{2}|\mathcal{W}'|^2}{n}[\frac{3}{N}+\frac{3}{n} + (\frac{2}{N^2}+\frac{2}{n^2})\mu_{w}\mu_{v}^2](\psi_{w,v}^{-})^2 < \sum_{w,v \in \mathcal{W}'}[\frac{24c^{2}k^{2}}{L^{2}T^{2}} + \mathcal{O}(\frac{1}{n})](\psi_{w,v}^{-})^2.
\end{aligned}
\end{equation}
Therefore, we have the second-order moment of $\Delta\mathcal{L}_{{\rm DI}}(\theta)$ decreases in the order of $\mathcal{O}(1)$:
\begin{equation}\label{eq:2Estimation}
\begin{aligned}
\centering
{\lim\limits_{n\to\infty}}\mathbb{E}[\Delta\mathcal{L}_{\rm DI}^2(\theta)] = \mathcal{O}(1),
\end{aligned}
\end{equation}
and thus converges to constant influenced by $\sum_{w,v \in \mathcal{W}'}(\psi_{w,v}^{-})^2$ in the limit of infinity.

\subsection{Complexity Analysis}\label{sec:complexity}
The computational cost of each operation in dynamic embedding model~(\ref{eq:DSGNS}) is the same as that of model~(\ref{eq:SGNS}).
Thus, the total computational cost is $\mathcal{O}((\Delta \mathcal{W}_{inc} + \Delta \mathcal{W}_{dis})k)$, where $k$ is the number of the negative samples.
In practice, we can use the size of the affected sub-graphs to evaluate the main computation complexity of network embedding learning.
According to the random walk and sliding window, the size of the affected nodes is $n_{add} + n_{van}$.
Therefore, the computation complexity of our dynamic network embedding framework is bounded by $\mathcal{O}((n_{add} + n_{van})k)$.
Similarly, the memory cost is bounded by $\mathcal{O}(n + n_{add} + n_{van})$.
Note that the memory cost of non-negative matrix factorization based temporal latent space network analysis approach \cite{Zhu2016Scalable} linearly grows over time.

\vspace{-0.15in}
\section{Experiments}\label{sec:experiments}\vspace{-0.05in}
We apply ISGNS to various large-scale real-world dynamic networks including a language network, three social networks and two citation networks (Section~\ref{sec:datasets}).
We empirically evaluate the time efficiency (Section~\ref{sec:timg_speedup}), the theoretical reliability (Section~\ref{sec:theorems}) and the quality of network representation (Section~\ref{sec:quality}) of the proposed ISGNS.

\subsection{Datasets}\label{sec:datasets}
The datasets used in this paper are Wikipedia, BlogCatalog, Flickr, Facebook, ArXiv, and DBLP networks.
A summarization of the statistics of the six datasets is shown in Table~\ref{tab:statistic}.

\begin{table}[h]\caption{\label{tab:statistic}Statistics of the dynamic network datasets}\small
\centering
\begin{tabular}{|c|c|c|c|c|}
\toprule
Name & $|V|$ & $|E|$ & Label & Time step\\
\hline
Wikipedia & 1,985,098 & 1,000924,086 &  7 & 16 \\
BlogCatalog & 10,312 &333,983 &  39 & 20 \\
Flickr & 80,513 & 5,899,882 & 195  & 100\\
Facebook & 1,715,256 & 22,613,981 & - & 24 \\
ArXiv & 18,722 & 198,110 & 195 & 16\\
DBLP & 524,061 & 20,580,238 & 100 & 730\\
\bottomrule
\end{tabular}\vspace{-0.05in}
\end{table}

\begin{figure}[h]
\centering
\includegraphics[width=0.45\textwidth,height=0.18\textheight]{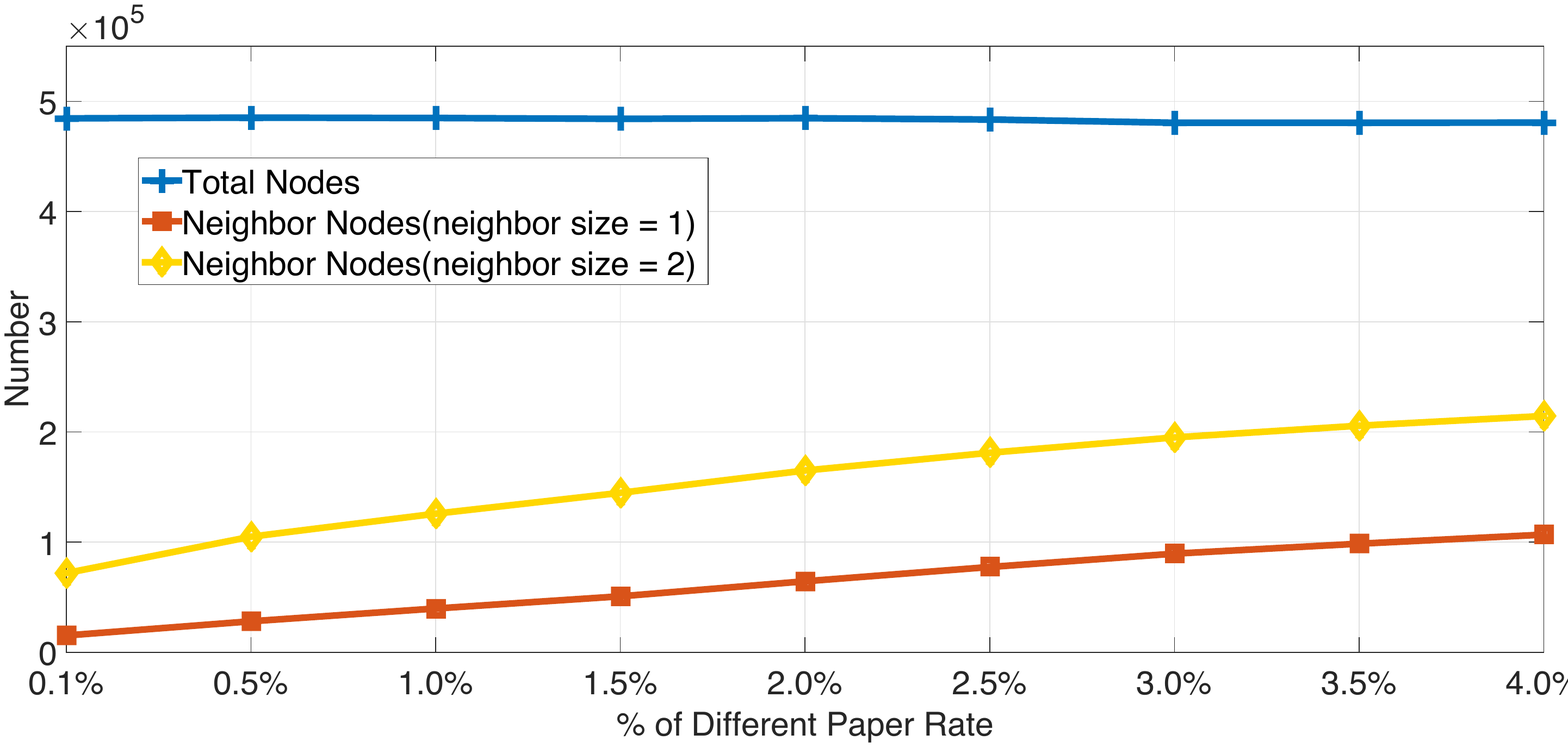}\vspace{-0.1in}
\caption{Number of the neighbor nodes with different settings}\centering\label{fig:neighbor_nodes}
\end{figure}

$\bullet$ \textbf{Wikipedia}:
This is a word co-occurrence network in the webpages of Wikipedia. We set an edge between two words if they co-occurr within the 5-words sliding window in the English Wikipedia pages.
The labels represent the Part-of-Speech (POS) tags inferred using the Stanford POS-Tagger.
In total the network has 1,985,098 nodes, 1,000,924,086 edges, and 7 different labels from the year 2001 to 2016.

$\bullet$ \textbf{BlogCatalog}~\cite{RezaSocial}:
This is a social blog directory which manages the bloggers and their blogs. It contains 10,312 bloggers as nodes and 333,983 relationships as edges in 20 days.
The labels represent the topic interests provided by the bloggers.
The network has 39 labels and a blogger may have multiple labels.

$\bullet$ \textbf{Flickr}~\cite{Tang2009Relational}:
This dataset is constructed with the images and the links among them collected from Flickr in 100 days.
The links between images represent they share common meta-data. In this data, edges are formed between two images that are taken from the same location.
This data has 80,513 nodes and 5,899,882 edges.

$\bullet$ \textbf{Facebook}~\cite{Mcauley2012Learning}: This is a social network dataset.
Nodes represent users, and edges are friendship relation between them.
The network has 4,039 nodes and 20,580,238 edges.

$\bullet$ \textbf{Arxiv}~\cite{Leskovec2007Graph}:
This is a collaboration network generated from the e-print arXiv and covers scientific collaborations between authors. 
Nodes represent scholars, and an edge represents two scholars having collaborated in a paper.
The network has 18,722 nodes and 198,110 edges.

$\bullet$ \textbf{DBLP}~\cite{Yang2012Defining}:
This is also a co-author network.
Nodes represent authors, and edges represent the co-author relation among them.
The network has 524,061 nodes and 970,742 edges.

\subsection{Training time and Speedup}\label{sec:timg_speedup}
We use the DBLP dataset to evaluate the training time and speedup performance of ISGNS.
In order to simulate the addition and deletion of nodes and edges, we build the dynamic co-author network.
We use 200,000 papers published earlier as the initial network, which contains 484,095 nodes and 970,742 edges.
Then we choose different sliding windows to move the the initial network along papers' publishing time. %by a variational sliding window of time.
More specific, we choose 0.10\%, 0.50\%, 1.0\%, 1.50\%, 2.00\%, 2.50\%, 3.00\%, 3.50\% and 4.00\% of time-series of window sliding rate, respectively, to add and delete related vertexes and edges to update the initial network as dynamic network.
Note that we consider both the addition and deletion of vertices and edges as shown in Table~\ref{tab:statistic_dy}.
We apply ISGNS to DeepWalk and node2vec models on the initial DBLP network with 200,000 papers, and run our algorithms~(\ref{eq:parametersinheritance1})-(\ref{eq:objective-frac2}) to update the noise distributions, the node vectors and the corresponding parameter vectors.
For comparisons, we also re-train the network embedding and run SGNS for the updated new network.
In the experiments, we run with 10 CPU threads, and the dimension of the generate network embedding is set to 128.

\begin{table}[h] \small
\caption{\label{tab:statistic_dy}The evolving procedure of the dynamic DBLP network}
\centering
\begin{tabular}{|c|c|c|c|c|c|}
\toprule
window sliding rate& 0.10\%& 0.50\%& 1.0\%& 1.50\%& 2.00\%\\
\hline
edge(+) & 2018& 4547& 7266& 9937& 13498\\
edge(-) & 1527& 3381& 6464& 9884& 12860\\
\hline
window sliding rate&  2.50\%& 3.00\%& 3.50\%& 4.00\% & - \\
\hline
edge(+) & 17054& 22842& 26340& 29217 & - \\
edge(-) & 16515& 19353& 22852& 25860 & - \\
\bottomrule
\end{tabular}\vspace{-0.05in}
\end{table}

\begin{figure*}[h]
    \subfloat[Training Time in DeepWalk]{\label{fig:time_deepwalk}\centering
\includegraphics[width=0.45\textwidth,height=0.18\textheight]{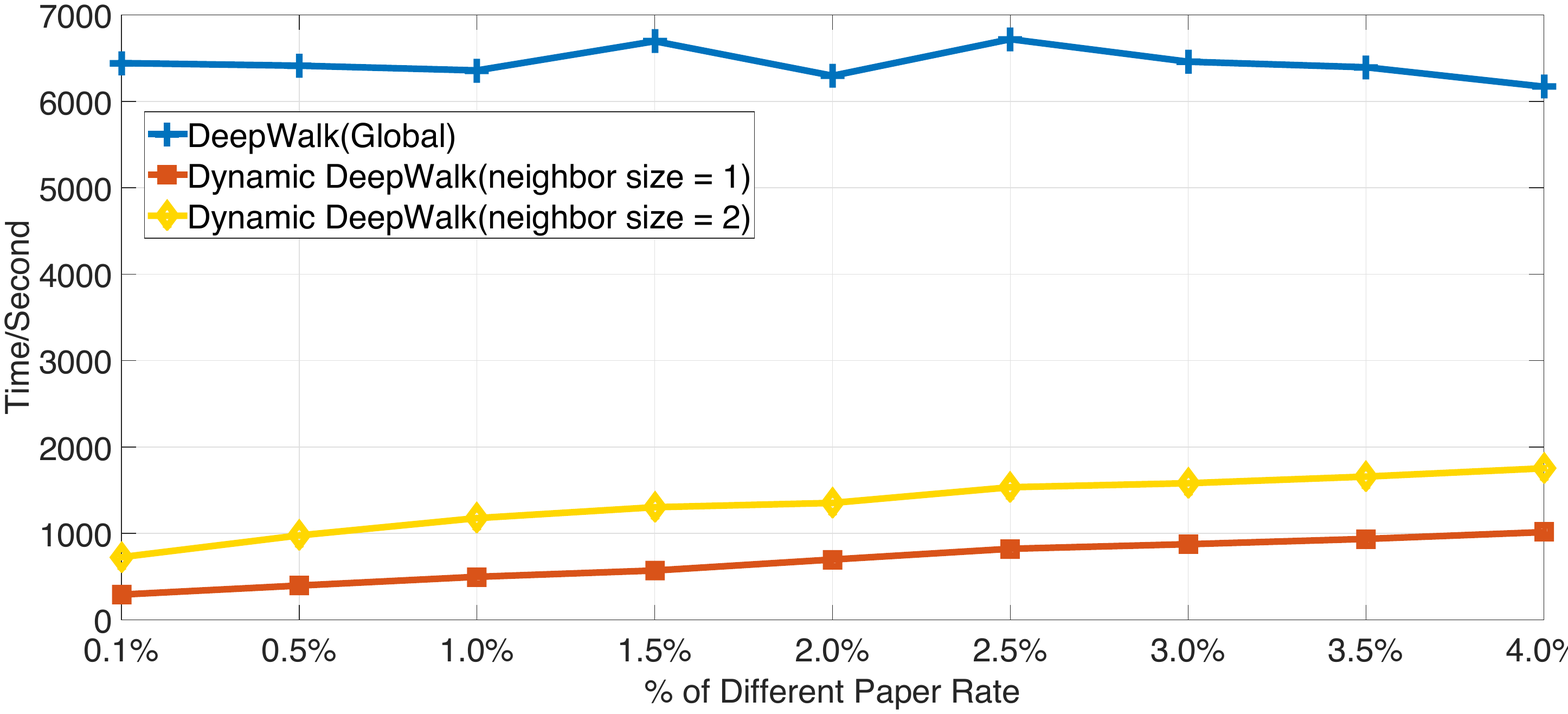}
	}
    \subfloat[Training Time in node2vec]{\label{fig:time_node2vec}\centering
\includegraphics[width=0.45\textwidth,height=0.18\textheight]{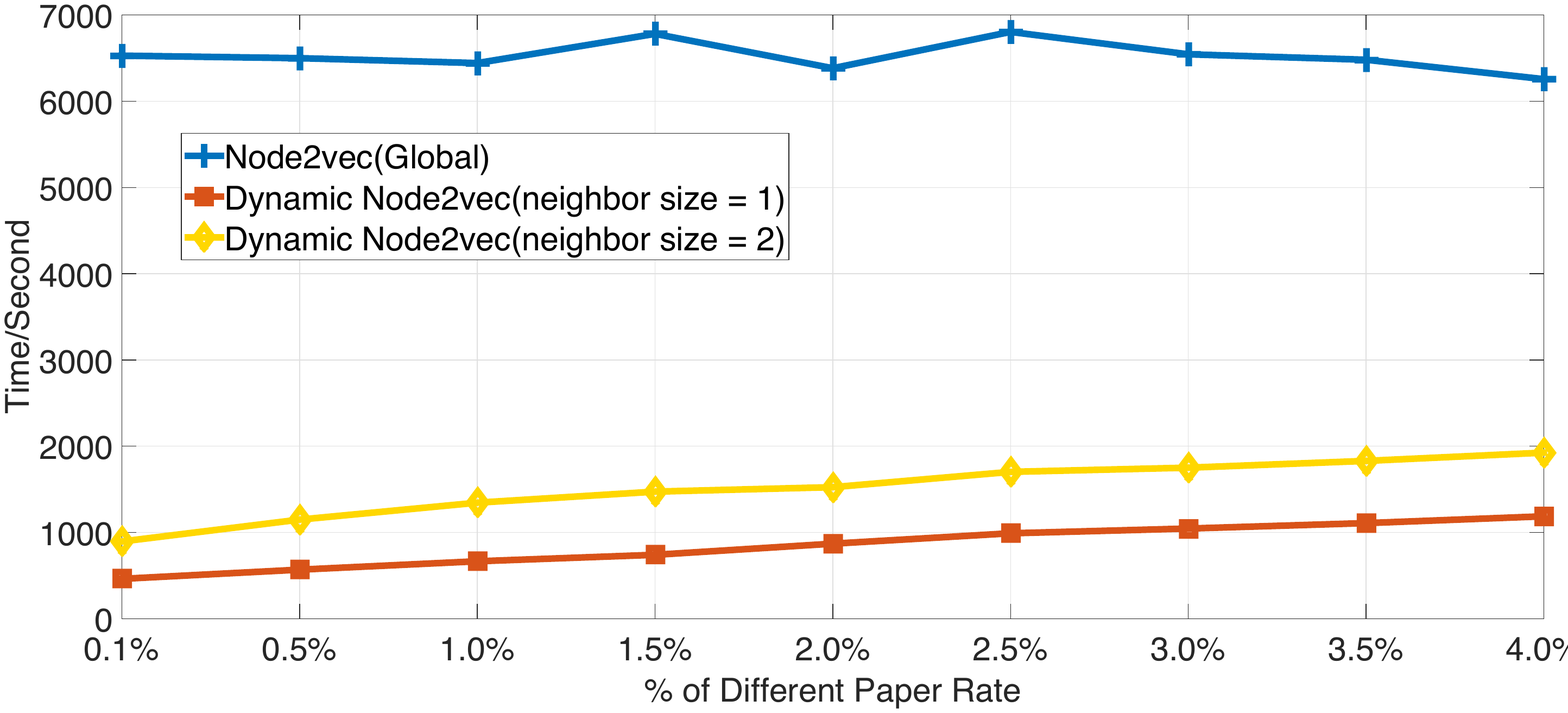}
	}\vspace{-0.1in}
	\caption{Training time of different methods under different settings.}\centering \label{fig:Vertex-time}
\end{figure*}

\begin{figure*}[h]\label{fig:speedup}
	\center
	\subfloat[Speedup in dynamic DeepWalk]{\label{fig:sp-deepwalk}
\includegraphics[width=0.45\textwidth,height=0.18\textheight]{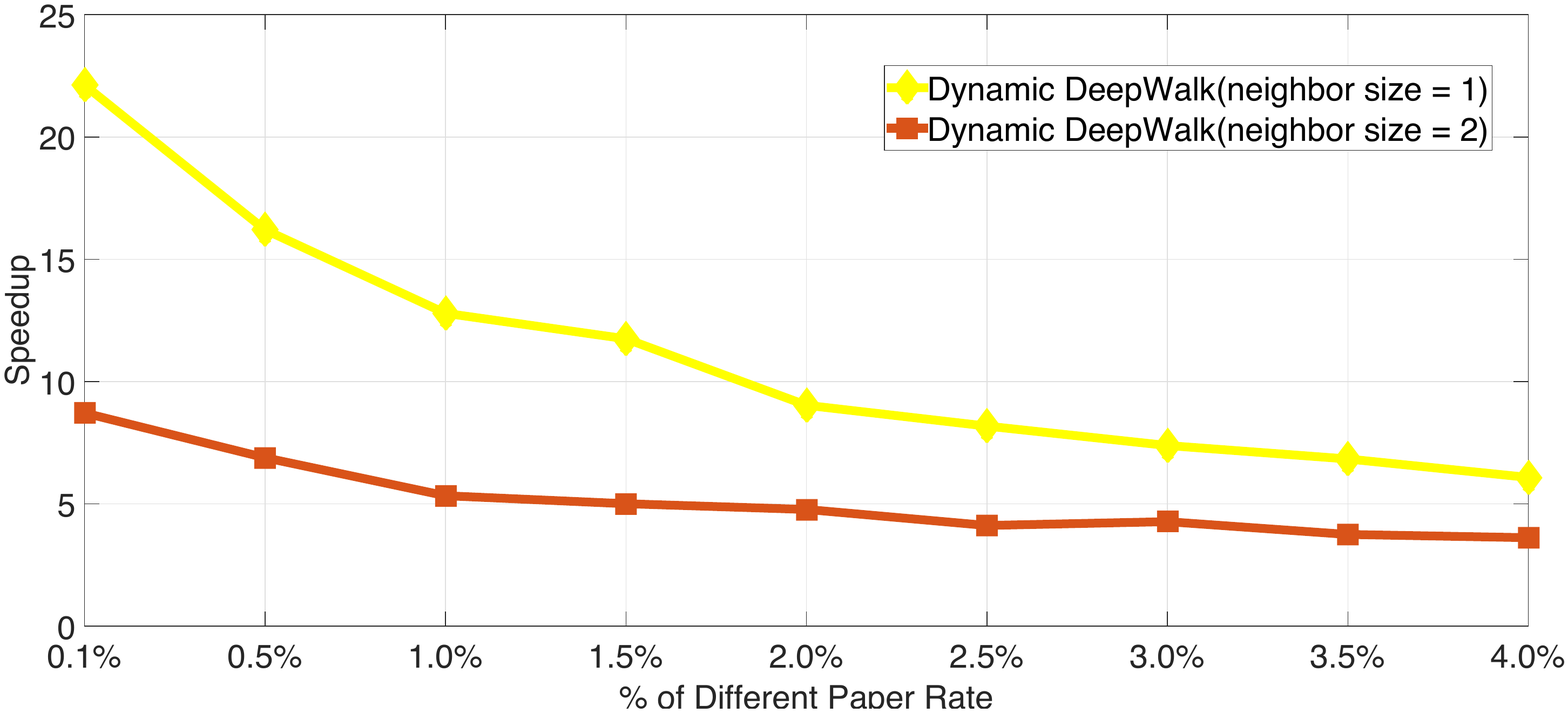}
	}
	\subfloat[Speedup in dynamic node2vec]{\label{fig:sp-node2vec}
\includegraphics[width=0.45\textwidth,height=0.18\textheight]{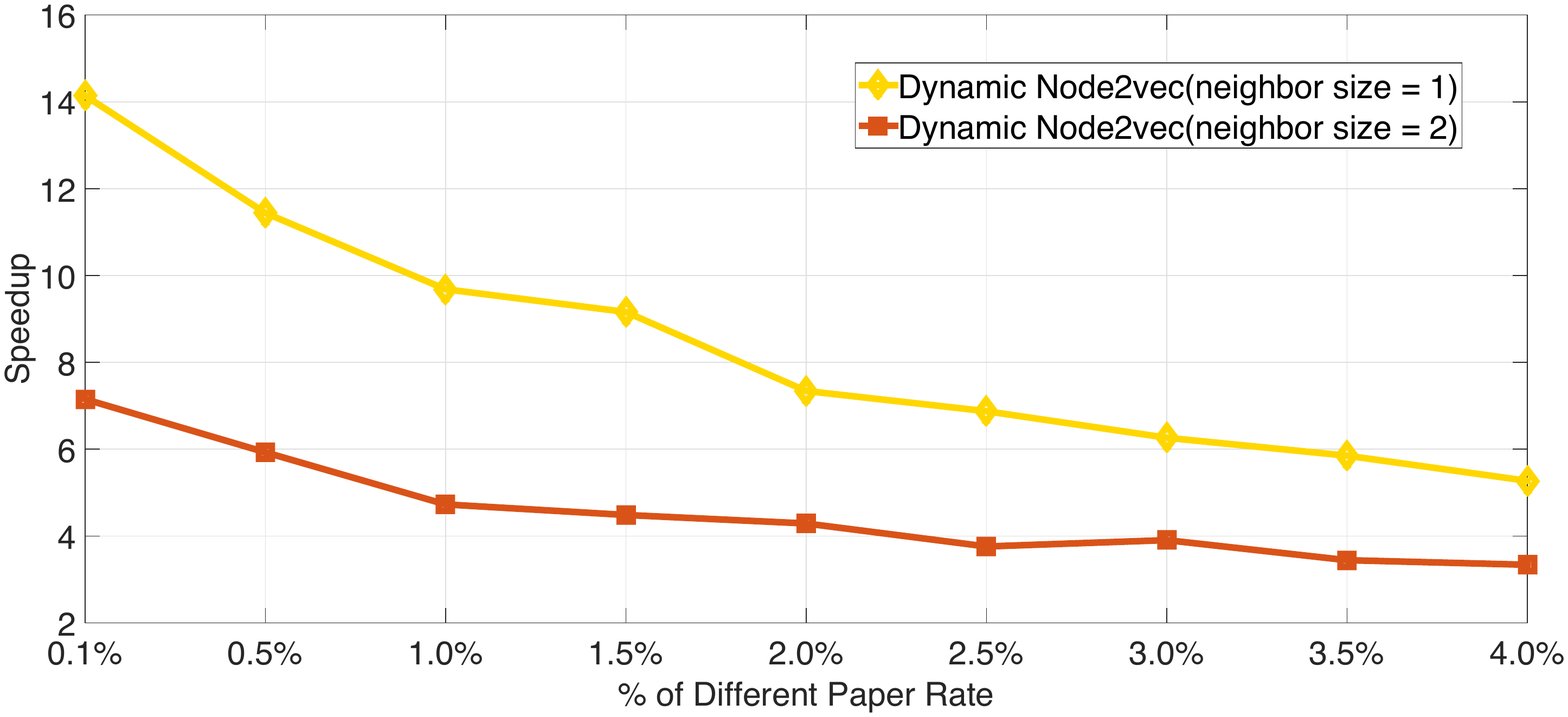}
	}\vspace{-0.1in}
	\caption{Speedup performances in dynamic DeepWalk and dynamic node2vec with different settings.}
\end{figure*}

We first check the updating network structure including the added and vanished vertices and edges by comparing the generated sequences of vertices before and after the random walk.
We retain the vectors related to the old vertices and update the sequences of vertices for more convenient gradient iterations in Eq.~(\ref{eq:objective-frac2}).
For the sequence $n$, if there are sliding windows of vertices containing the newly vanished nodes, we update the vectors of the vertices in the sliding windows with stochastic gradient descent method.
For the sequence $N$, if there are sliding windows of vertices containing the newly added nodes, we update the vectors of the vertices in the sliding windows with stochastic gradient ascent method. We count the rates of the nodes in the affected neighbor sub-graphs, as shown in Figure~\ref{fig:neighbor_nodes}.
The blue line refers to the total number of nodes in the dynamic scenario.
Note that the total number of nodes is changing.
The orange and yellow lines are the number of the affected nodes when the sliding window arises from 3 to 5.

We apply ISGNS to DeepWalk and node2vec, and check the training time and the achieved speedup. The results are shown in Figure~\ref{fig:time_deepwalk} and Figure~\ref{fig:time_node2vec}.
It is shown that the time consumption of ISGNS linearly increases with the increase of the size of the updated nodes and edges in the dynamic network.
Although the sliding rate from 0.10\% to 4.00\% is relatively small compared to the original network, the proportion of affected neighbor nodes is relatively large rising from about 14.83\% to 44.60\% due to the dense connectivity among the nodes.
We employ stochastic gradient method to update the vectors among sliding windows of the sequence of vertices following Eq.~(\ref{eq:objective-frac2}).
The time consumptions of the batch global network embedding models depend on the overall scale of the networks.
It costs about 6500 and 6800 seconds when the size of the nodes is 484000, while our models take much less time than batch global re-training.
One can also see that the time consumptions in DeepWalk and node2vec both linearly increase with the change rate increasing for dynamic networks.

The speedup results are shown in Figure~\ref{fig:sp-deepwalk} and Figure~\ref{fig:sp-node2vec}.
One can see that for the smaller change in the dynamic network, the speedup is more significant.
DeepWalk with ISGNS achieves up to 22 times speedup, while  node2vec with ISGNS has up to 14 times speedup.

\subsection{Validation of Theoretical Analysis}\label{sec:theorems}
Now we give an empirical experiment to validate our theoretical analysis in Section 4.
Since it is difficult to assess the node vector value generated by stochastic gradient optimization models between Eq.~(\ref{eq:DSGNS}) and Eq.~(\ref{eq:SGNS}) directly, we focus on verifying the boundness analysis and give the limit of first-order and second-order moments of objective difference.
As we proved the limits of infinity in Eq.~(\ref{eq:Estimation-Euler-Harmonic1}) and Eq.~(\ref{eq:2Estimation}), the first-order and second-order moments of objective difference are affected by the old sequence of vertices $n$.
We measure the first and second order of moments on Facebook dataset with various network sizes $\mathbb{W}$ varying over \{$10^2, 2\times10^{2}, 2^{2}\times10^2, 2^{3}\times10^2, \cdots, 2^{14}\times10^2$\}.
We also choose $ 1\%, 5\%, 10\%, 15\%$ of the network nodes change rates for the above dynamic Facebook network.
Since the second-order moment can be affected by the  times and length of random walk for each node.
For all the experiments, we set the length and times of random walk as 80 and 100, respectively.

\begin{figure*}[h]\label{fig:speedup}
	\center
    \subfloat[First-order moment]{\label{fig:first_order_moment}
\includegraphics[width=0.45\textwidth,height=0.18\textheight]{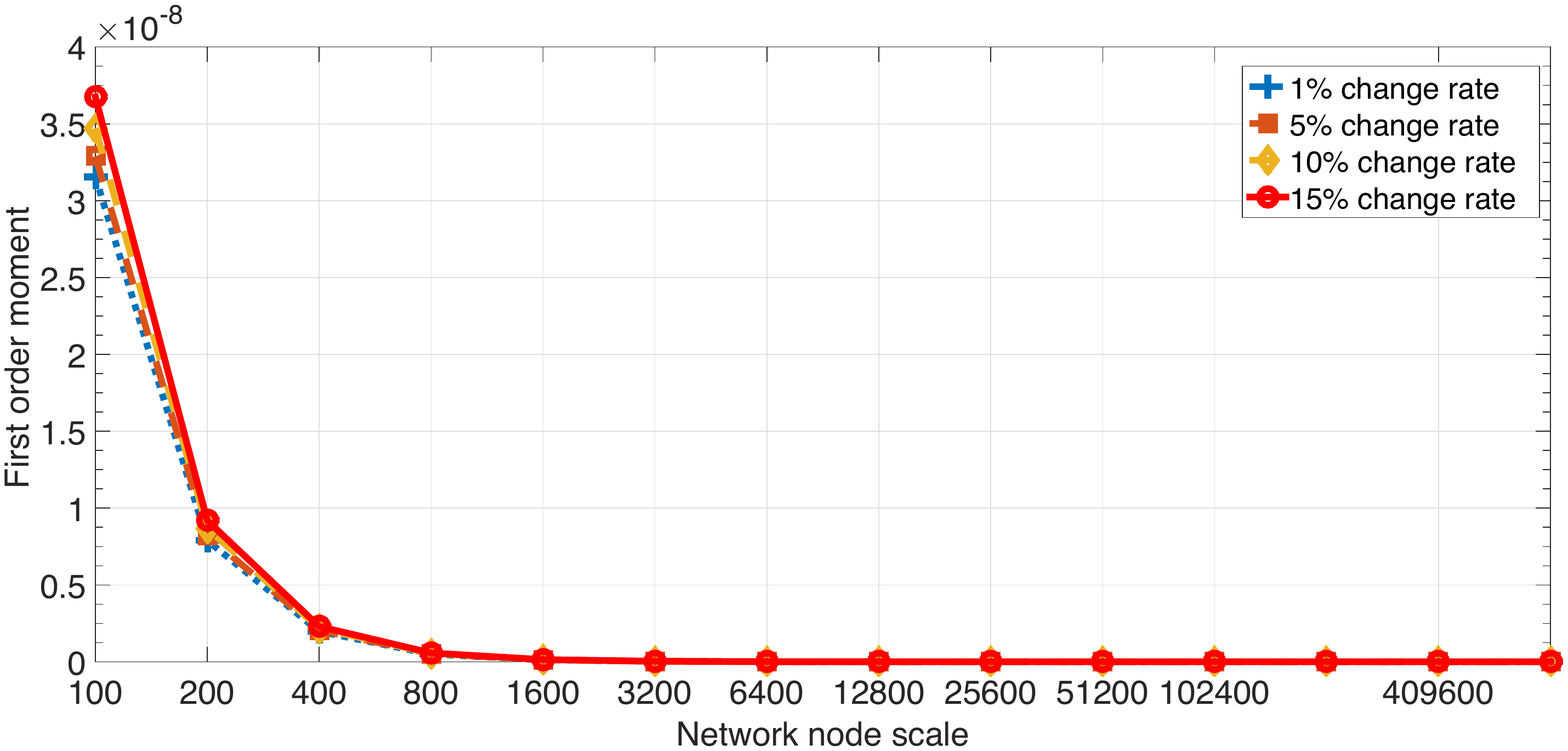}
	}
	\subfloat[Second-order moment]{\label{fig:second_order_moment}
\includegraphics[width=0.45\textwidth,height=0.18\textheight]{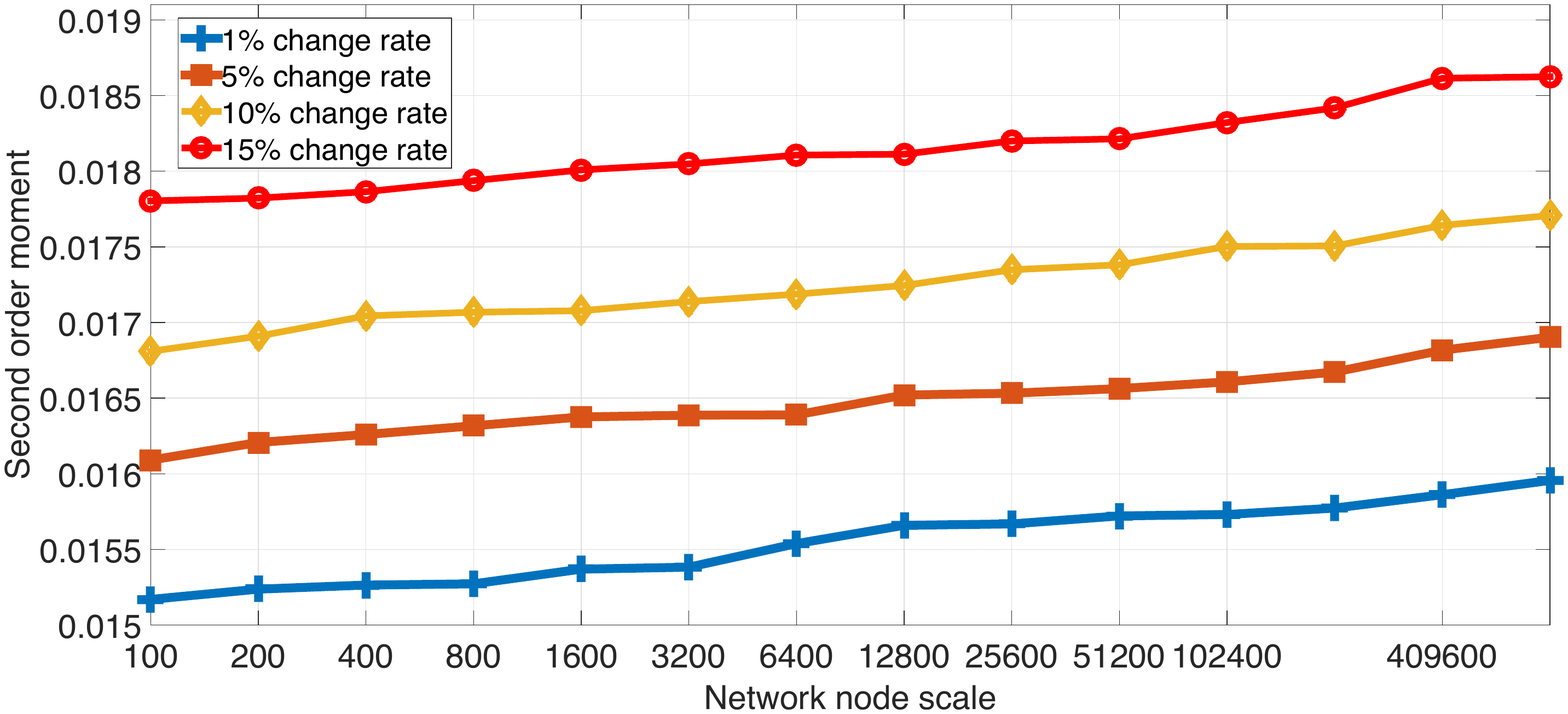}
	}\vspace{-0.1in}
	\caption{First-order moment and second-order moment with different change rates.}
\end{figure*}

Figure~\ref{fig:first_order_moment} shows the first-order moment of $\Delta\mathcal{L}_{DI}(\theta)$ computed on the different sizes of the training data and different network change rates.
Since the Eq.~(\ref{eq:Estimation-Euler-Harmonic1}) suggests that the first-order moment decreases in the order of $\mathcal{O}(\frac{1}{n^2})$.
The expectation of $\mathbb{E}[\Delta\mathcal{L}_{DI}(\theta)]$ converges to zero when the network size tends to be infinitely great.
In Figure~\ref{fig:first_order_moment}, the four lines are close to each other, which demonstrates the smaller the change rate of the network, the smaller the expectation of objective difference.
Note that the x-axis is log scale, and the first-order moments of magnitudes decrease from $10^{-8}$ to $10^{-16}$.

Figure~\ref{fig:second_order_moment} shows the second-order moment of $\Delta\mathcal{L}_{DI}(\theta)$ computed on different sizes of datasets and different network change rates.
Different from the first-order moments, We can see that the second order moments slowly increases with the same exponentially growing network scales.
However, the limits of infinity of $\mathbb{E}[\Delta\mathcal{L}_{\rm DI}^2(\theta)]$ is $\mathcal{O}(1)$.
Similar to the trend in first-order moments, a smaller change rate of network leads to a smaller second order moments of the objective difference.
Although the second order moments do not converge to zero, the real values are small and change in a relatively small range from $1.52\times 10^{-2}$ to $1.87\times 10^{-2}$.

\subsection{Quality of Network Embeddings}\label{sec:quality}
This experiment aims to investigate the quality of the network embeddings learned by our dynamic network embeddings through comparison with the batch global re-training based counterparts and other dynamic network embedding approach~\cite{Li2017Attributed}.
We employ the multi-label classification and link predication tasks~\cite{Grover2016node2vec,Perozzi2014DeepWalk,Li2017Attributed} to evaluate the quality of the learned network embeddings.

\begin{table*}[h]\caption{\label{tab:multi_label_class_deepwalk}Multi-label classification results by DeepWalk}\small
\centering
\begin{tabular}{|c|c|c|c|c|c|c|c|c|c|c|c|}
\toprule
Item & DS & Metric & 10\%& 20\%& 30\%& 40\%& 50\%& 60\%& 70\%& 80\%& 90\%\\
\hline
Dy &BC & Mi-F1& \textbf{36.02\%} & \textbf{36.21\%} & \textbf{39.61\%} & 40.28\% & \textbf{41.11\%} & 41.29\% & \textbf{41.51\%} & 41.47\% & \textbf{42.05\%} \\
Gl &BC & Mi-F1&  36.00\% & 36.20\% & 39.60\% & \textbf{40.30\%} & 41.00\% & \textbf{41.30\%} & 41.50\% & \textbf{41.50\%} & 42.00\% \\
Dy &BC & Ma-F1& \textbf{21.31\%} & \textbf{23.81\%} & \textbf{25.31\%} & 26.29\% & 27.33\% & \textbf{27.60\%} & \textbf{27.90\%} & 28.18\% & \textbf{28.92\%} \\
Gl &BC & Ma-F1&  21.30\% & 23.80\% & 25.30\% & \textbf{26.30\%} & \textbf{27.30\%} & \textbf{27.60\%} & \textbf{27.90\%} & \textbf{28.20\%} & 28.90\% \\
Dy &Fl & Mi-F1& 32.30\% & 34.59\% & 36.11\% & \textbf{36.88\%} & \textbf{37.21\%} & 37.77\% & 38.05\% & \textbf{38.43\%} & \textbf{38.88\%} \\
Gl &Fl & Mi-F1& \textbf{32.44\%} & \textbf{34.61\%} & 35.94\% & 36.79\% & \textbf{37.21\%} & \textbf{37.79\%} & \textbf{38.13\%} & 38.41\% & 38.76\% \\
Dy &Fl & Ma-F1& 13,91\% &17.14\% & \textbf{19.74\%} & \textbf{21.16\%} & \textbf{22.07\%} & 22.75\% & 23.55\% & \textbf{24.11\%} & \textbf{24.78\%} \\
Gl &Fl & Ma-F1& \textbf{14.06\%} & \textbf{17.17\%} & 19.69\% & 21.11\% & 22.05\% & \textbf{22.78\%} & \textbf{23.62\%} & 24.10\% & 24.72\% \\
Dy &Wp & Mi-F1& \textbf{78.87\%} & 79.91\% & \textbf{80.42\%} & \textbf{80.72\%} & \textbf{80.93\%} & 81.15\% & \textbf{81.27\%} & 81.33\% & 81.42\% \\
Gl &Wp & Mi-F1& 78.86\% & \textbf{79.93\%} & 80.41\% & 80.69\% &\textbf{ 80.93\%} & \textbf{81.16\%} & 81.25\% & \textbf{81.35\%} & \textbf{81.43\%} \\
Dy &Wp & Ma-F1& \textbf{78.72\%} &79.74\% & \textbf{80.33\%} & \textbf{80.56\%} & \textbf{80.81\%} & 80.93\% & \textbf{81.11\%} & 81.21\% & 81.21\% \\
Gl &Wp & Ma-F1& 78.71\% & \textbf{79.76\%} & 80.32\% & 80.50\% & \textbf{80.81\%} & \textbf{80.94\%} & 81.10\% & \textbf{81.23\%} & \textbf{81.31\%} \\
\bottomrule
\end{tabular}\vspace{-0.05in}
\end{table*}

\begin{table*}[h]\caption{\label{tab:multi_label_class_node2vec}Multi-label classification results by node2vec}\small
\centering
\begin{tabular}{|c|c|c|c|c|c|c|c|c|c|c|c|}
\toprule
Item & DS & Metric & 10\%& 20\%& 30\%& 40\%& 50\%& 60\%& 70\%& 80\%& 90\%\\
\hline
Dy &BC & Mi-F1& \textbf{36.71\%} & \textbf{37.19\%} & \textbf{39.99\%} & \textbf{40.30\%} & \textbf{41.29\%} & \textbf{42.06\%} & 41.44\% & 42.57\% & \textbf{42.87\%} \\
Gl &BC & Mi-F1&  36.70\% & 37.17\% & 39.98\% & \textbf{40.30\%} & 41.27\% & \textbf{42.06\%} & \textbf{41.46\%} & \textbf{42.58\%} & 42.86\% \\
Dy &BC & Ma-F1& \textbf{21.40\%} & \textbf{23.97\%} & \textbf{25.37\%} & \textbf{26.39\%} & \textbf{27.51\%} & 27.69\% & 27.96\% & \textbf{28.21\%} & \textbf{28.97\%} \\
Gl &BC & Ma-F1&  \textbf{21.40\%} & 23.96\% & \textbf{25.37\%} & 26.38\% & 27.50\% & \textbf{27.70\%} & \textbf{27.97\%} & \textbf{28.21\%} & 28.96\% \\
Dy &Fl & Mi-F1& \textbf{33.59\%} & 35.15\% & \textbf{37.11\%} & \textbf{37.93\%} & 38.26\% & \textbf{38.91\%} & \textbf{38.99\%} & 39.14\% & \textbf{39.44\%} \\
Gl &Fl & Mi-F1& 33.57\% & \textbf{35.16\%} & 36.96\% & 37.84\% & \textbf{38.27\%} & 38.90\% & 38.95\% & \textbf{39.17\%} & 39.42\% \\
Dy &Fl & Ma-F1& 14.11\% & \textbf{18.21\%} & 20.41\% & \textbf{22.25\%} & 23.27\% & 23.26\% & \textbf{24.72\%} & \textbf{25.81\%} & 25.91\% \\
Gl &Fl & Ma-F1& \textbf{14.12\%} & 18.18\% & \textbf{20.43\%} & 22.24\% & \textbf{23.30\%} & \textbf{23.28\%} & 24.68\% & 25.79\% & \textbf{25.94\%} \\
Dy &Wp & Mi-F1& \textbf{79.05\%} & \textbf{80.05\%} & \textbf{80.75\%} & \textbf{80.89\%} & \textbf{81.39\%} & \textbf{81.33\%} & \textbf{81.55\%} & 81.62\% & \textbf{81.69\%} \\
Gl &Wp & Mi-F1& 79.04\% & \textbf{80.05\%} & 80.74\% & 80.87\% & 81.38\% & 81.31\% & \textbf{81.55\%} & \textbf{81.63\%} & \textbf{81.69\%} \\
Dy &Wp & Ma-F1& \textbf{78.97\%} & 79.82\% & \textbf{80.60\%} & \textbf{80.71\%} & \textbf{81.28\%} & \textbf{80.26\%} & \textbf{81.11\%} & 81.47\% & \textbf{81.57\%} \\
Gl &Wp & Ma-F1& 78.96\% & \textbf{79.83\%} & 80.59\% & 80.70\% & 81.27\% & 80.25\% & 81.10\% & \textbf{81.48\%} & 81.56\% \\
\bottomrule
\end{tabular}\vspace{-0.05in}
\end{table*}

\begin{table}[h]\caption{\label{tab:comparsion-multi}Multi-label classification results comparison of different embedding methods}\small
\centering
\begin{tabular}{|c|c|c|c|c|c|c|}
\toprule
 Dataset& Algorithms & Micro-F1& Macro-F1\\
 \hline
 BC & Dynamic DeepWalk & 42.05\% & 28.92\%  \\
 BC & Dynamic node2vec & \textbf{42.87\%} & \textbf{28.97\%}  \\
 BC & DANE\cite{Li2017Attributed} & 43.27\%  & 29.12\% \\
 BC & Dynamic SBM\cite{Xu2013Dynamic} & 39.41\% & 22.63\% \\
 BC & DNE\cite{du2018dynamic} & 40.75\% & 26.13\% \\
 Fl & Dynamic DeepWalk & 38.88\% & 24.78\% \\
 Fl & Dynamic node2vec & \textbf{39.44\%} & \textbf{25.91\%}  \\
 Fl & DANE\cite{Li2017Attributed} & 32.81\%  & 20.75\% \\
 Fl & Dynamic SBM\cite{Xu2013Dynamic} & 36.52\% & 23.87\% \\
 Fl & DNE\cite{du2018dynamic} & 37.87\% & 24.28\% \\
\bottomrule
\end{tabular}\vspace{-0.05in}
\end{table}

\subsubsection{Multi-label Classification}
We firstly evaluate the quality of the learned dynamic network embeddings through the multi-label classification task on the Wikipedia (Wp), BlogCatalog (BC) and Flickr (Fl) datasets.
In order to construct a dynamic network, we take full advantage of the time attributes of nodes and edges on the three networks.
We adopt two kinds of training modes, including dynamic training (Dy) over time and batch global training (Gl), to extract the dynamic network features.
For the dynamic training, we adopt the time sequence batch network structures.
BlogCatalog, Flickr and Wikipedia are the time-series accumulative social relationship networks, as shown in Table~\ref{tab:statistic}.
We use an online learning method to train the three dynamic network datasets over time.
However, in addition to the dynamic time factor, we also inherit the original random walk strategies and the dimension of the vector configurations.
We inherit the parameter pairs $\{(0.25,0.25), (0.25,0.25), (4,0.5)\}$ for BlogCatalog, Flickr and Wikipedia from node2vec\cite{Grover2016node2vec}.

We train the node feature representations, including dynamic network embeddings considering time factor in Eq.~(\ref{eq:DSGNS}) and one-round global network embeddings without consideration of time factor in Eq.~(\ref{eq:SGNS}), as the input to a one-vs-rest logistic regression classifier with L2 regularization implemented by LibLinear~\cite{Fan2008LIBLINEAR}.
Specifically, we randomly sample a portion of the labeled nodes and use them as training data.
The rest of the nodes are used as the test data.
We also randomly and equally split the train and test data over 10 parts, perform a 10-fold cross validation, and report the average Micro-F1 and Macro-F1.
From the results given in Table~\ref{tab:multi_label_class_deepwalk} and Table~\ref{tab:multi_label_class_node2vec}, one can see that the dynamic network embedding results are comparable to and sometimes better than the global training results.
Both the Micro-F1 and Macro-F1 are reported when the labeled node percentage increases from 10\% to 90\%.
Overall, the dynamic network embedding performs better than one round of global training.
This may be because some important dynamic structure patterns are sufficiently trained and captured.
Moreover, the node2vec based embedding models perform better than DeepWalk in experiments.

We further compare with the state-of-the-art dynamic network embedding methods including DANE~\cite{Li2017Attributed}, Dynamic SBM~\cite{Xu2013Dynamic} and DNE~\cite{du2018dynamic} on the multi-label classification task.
Since the vertices of our networks do not have attribute information, we adopt the DANE with only network information for fairness.
As shown in Table~\ref{tab:comparsion-multi}, the neural network based network embeddings are better than spectral embedding based DANE and statistical model based Dynamic SBM models
The experiments demonstrate the generality of ISGNS model.
The experimental results also show that the convergent ISGNS has better learning ability for dynamic network embedding than the heuristic models.

\subsubsection{Link Prediction}
In the link prediction task, we are given a network with a certain fraction of missing edges, and we need to predict the missing edges.
To facilitate the comparison between the dynamic embedding method and the baselines, we use the same datasets and experiment setting as in~\cite{Grover2016node2vec}.
We generate the labeled dataset of edges as follows.
We randomly remove 50\% of edges from the network as the positive samples.
We randomly sample an equal number of node pairs from the network which actually have no edges connecting them as the negative examples.
We conduct the experiment on Facebook (Fb) and Arxiv(Ax) datasets, and use the Area Under Curve (AUC) scores for link prediction with four different binary operators, including Average, Hadamard, Weighted-L1 and Weight-L2, for learning edge features~\cite{Grover2016node2vec}.

\begin{table*}[h]\caption{\label{tab:link_prediction}Area Under Curve (AUC) scores for link prediction}\small
\centering
\begin{tabular}{|c|c|c|c|c|c|}
\toprule
Dataset& Algorithm & Average & Hadamard & Weighted-L1 & Weighted-L2 \\
\hline
Fb& Dynamic DeepWalk & \textbf{0.7268} & \textbf{0.9548} & \textbf{0.9474} & \textbf{0.9536} \\
Fb& Global DeepWalk & 0.7261 & 0.9544 & 0.9461 & 0.9535 \\
Fb& Dynamic node2vec & 0.7266 & 0.9555 & 0.9504 & 0.9526 \\
Fb& Global node2vec & 0.7264 & 0.9554 & 0.9503 & 0.9524\\
Ax & Dynamic DeepWalk &0.7058 &0.9275 &0.8186 &0.8278 \\
Ax & Global Deepwalk & 0.7056 & 0.9274 &  0.8183 & 0.8276\\
Ax &Dynamic node2vec & \textbf{0.7204} & \textbf{0.9305} & \textbf{0.8371} & \textbf{0.8474} \\
Ax &Global node2vec & 0.7203 & 0.9305 & 0.8371 & 0.8474 \\
\bottomrule
\end{tabular}\vspace{-0.05in}
\end{table*}

From the results in Table~\ref{tab:link_prediction}, one can see that similar to the multi-label classification task, the performances of dynamic network embedding in link prediction are comparable with and sometimes better than the global training results on the two datasets.
The four binary operators which generate edge features are reported from dynamic embeddings to global embeddings.
On the whole, the dynamic embeddings results are better than one round of global training for networks of sequential growth.
This again demonstrates the generality of our dynamic skip-gram with negative sampling framework.

\vspace{-0.15in}
\section{Conclusion}\label{sec:conclusion}\vspace{-0.05in}
This paper proposed a dynamic network embedding framework based on the incremental skip-gram with negative sampling from both practical and theoretical perspectives.
Theoretical analysis showed that the objective difference can be bounded by a function of the number of changed nodes and links, and the first-order moment of objective difference can be convergent in order of $\mathbb{O}(\frac{1}{n^2})$, and the second-order moment of objective difference can be stabilized in order of $\mathbb{O}(1)$.
The results of the systematic evaluations on multi-label classification task and link prediction tasks over multi real-world dynamic network datasets show that our dynamic network embedding framework is significantly faster than global training, and achieved comparable network embedding performance.
The success of this work proves the scalability and robustness of skip-gram with negative sampling algorithm.
A potential future work is to extend our approach to other advanced network representation learning models~\cite{Tang2015LINE,Cao2015GraRep,Wang2016Structural,Swami2017metapath2vec,Yang2017Fast,Goyal2018Graph}.

%%%%%%%%%%%%%%%%%%%%%%%%%%%%%%%%%%%%%%%%%%%%%%%%%%%%%%%
%%% Acknowledgements. ÖÂÐ»
%%%%%%%%%%%%%%%%%%%%%%%%%%%%%%%%%%%%%%%%%%%%%%%%%%%%%%%
\Acknowledgements{This work was supported by NSFC program through grants 61872022, 61772151, 61421003, National Key R\&D Program of China through grant 2017YFB0803305, and SKLSDE-2018ZX-16.}

\bibliography{sigproc}

%%%%%%%%%%%%%%%%%%%%%%%%%%%%%%%%%%%%%%%%%%%%%%%%%%%%%%%
%%% Supplements. ²¹³ä²ÄÁÏ, ·Ç±ØÑ¡
%%%%%%%%%%%%%%%%%%%%%%%%%%%%%%%%%%%%%%%%%%%%%%%%%%%%%%%
%\Supplements{Appendix A.}

%%%%%%%%%%%%%%%%%%%%%%%%%%%%%%%%%%%%%%%%%%%%%%%%%%%%%%%
%%% Reference section. ²Î¿¼ÎÄÏ×
%%% citation in the content using "some words~\cite{1,2}".
%%% ~ is needed to make the reference number is on the same line with the word before it.
%%%%%%%%%%%%%%%%%%%%%%%%%%%%%%%%%%%%%%%%%%%%%%%%%%%%%%%
%\begin{thebibliography}{99}

%\bibitem{1} Author A, Author B, Author C. Reference title. Journal, Year, Vol: Number or pages

%\bibitem{2} Author A, Author B, Author C, et al. Reference title. In: Proceedings of Conference, Place, Year. Number or pages

%\end{thebibliography}

%%%%%%%%%%%%%%%%%%%%%%%%%%%%%%%%%%%%%%%%%%%%%%%%%%%%%%%
%%% Appendix sections. ¸½Â¼ÕÂ½Ú, ·Ç±ØÑ¡
%%%%%%%%%%%%%%%%%%%%%%%%%%%%%%%%%%%%%%%%%%%%%%%%%%%%%%%
%\begin{appendix}
%\section{Name}

%\end{appendix}

\end{document}